\documentclass{comjnl}

\usepackage{amsmath}
\usepackage{amsopn}
\usepackage{times}
\usepackage{epsfig}
\usepackage{graphicx}
\usepackage{amssymb}
\usepackage{amstext}
\usepackage{booktabs}
\usepackage{bbm}
\usepackage{stfloats}
\usepackage{multicol}
\usepackage{gensymb}
\usepackage[basic]{mathastext}
\usepackage{multirow}
\usepackage{array, caption, threeparttable}
\usepackage{ulem}
\usepackage{amsbsy}
\usepackage{algorithm}  
\usepackage{algorithmic}
\usepackage{nicefrac}
\usepackage{amssymb}
\usepackage{pifont}
\newcommand{\xmark}{\ding{56}}
\usepackage{subfigure}
\usepackage[utf8]{inputenc}
\DeclareMathOperator*{\argmin}{arg\,min}

\begin{document}

\title[Improved Loss Function-Based Prediction Method of Extreme Temperatures in Greenhouses]{Improved Loss Function-Based Prediction Method of Extreme Temperatures in Greenhouses}
\author{Liao Qu}
\author{Shuaiqi Huang}
\author{Yunsong Jia}
\author{Xiang Li}
\affiliation{College of Information and Electrical Engineering, China Agricultural University, Beijing, China} 
\email{cqlixiang@cau.edu.cn}

\shortauthors{Liao Qu, Shuaiqi Huang, Yunsong Jia, Xiang Li}



\begin{abstract}
The prediction of extreme greenhouse temperatures to which crops are susceptible is essential in the field of greenhouse planting. It can help avoid heat or freezing damage and economic losses. Therefore, it’s important to develop models that can predict them accurately. Due to the lack of extreme temperature data in datasets, it is challenging for models to accurately predict it. In this paper, we propose an improved loss function, which is suitable for a variety of machine learning models. By increasing the weight of extreme temperature samples and reducing the possibility of misjudging extreme temperature as normal, the proposed loss function can enhance the prediction results in extreme situations. To verify the effectiveness of the proposed method, we implement the improved loss function in LightGBM, long short-term memory, and artificial neural network and conduct experiments on a real-world greenhouse dataset. The results show that the performance of models with the improved loss function is enhanced compared to the original models in extreme cases. The improved models can be used to guarantee the timely judgment of extreme temperatures in agricultural greenhouses, thereby preventing unnecessary losses caused by incorrect predictions.
\end{abstract}

\maketitle

\section{Introduction}

The greenhouse is a typical scenario of modern agriculture\cite{kang2018managing}. Among the various environmental factors, the temperature has the most significant impact on the growth of crops \cite{Hatfield2011ClimateIO,luo2011temperature}. Temperature influences many biochemical processes such as photosynthesis, respiration, water balance, and membrane stability of leaves \cite{Kaushal2016FoodCF}, which would then affect the quality and yields of crops. Therefore, accurate temperature prediction is the key to greenhouse management.
Although fluctuations within normal temperature do not affect the growth of crops, the extreme temperature may result in severe damage to them\cite{wang2003plant,hatfield2015temperature}. For example, under extremely high temperatures, crops will suffer thermal damage, which affects the process of pollination and fruiting, possibly leading to death \cite{Ferris1998EffectOH,siebert2014future}. Similarly, under extremely low temperatures, crops will freeze or suffer frostbite\cite{Huner2004PhotosynthesisPA, Liu2013TheLT}. Hence, accurate prediction of extreme temperatures is conducive to early prevention of such damage.
In recent years, based on computational descriptions of greenhouse environment \cite{Tong2009NumericalMO,Vanthoor2011AMF} and statistical methods \cite{Yu2016PredictionOT,Hongkang2018RecurrentNN,Luo2020ANF,Lins2013PredictionOS,Ahmad2014NeuralNM,Singh2017PREDICTIONOG}, various models have been successfully developed to predict the greenhouse temperature.  However, in contrast to mathematical models based on the internal biology mechanism of the crops, environment, and production process in the greenhouse, machine learning models do not require many physical factors \cite{Nayak2008EnergyAE}, they only need historical environment data of greenhouses. Because using machine learning in temperature prediction is relatively straightforward, many related studies have been conducted in recent years. \cite{Ahmad2014NeuralNM} used an artificial neural network(ANN) with the Levenberg-Marquardt method \cite{more1978levenberg} to predict temperature and humidity inside a naturally ventilated greenhouse, achieving error index of 0.0025 for temperature and 0.0024 for humidity. \cite{Singh2017PREDICTIONOG} collected several micro-climate parameters of a greenhouse including temperature, relative humidity, wind speed, and solar radiation located in a sub-humid sub-tropical region of India. They then explored the optimal parameters for artificial neural network models and made one-day-ahead predictions of air temperature and relative humidity of a greenhouse and achieved a mean absolute error of 0.558°C and 1.976\% in their dataset. \cite{Hongkang2018RecurrentNN} constructed a recurrent neural network using substrate temperature, air humidity, illumination, and concentration for model training and achieved results with root-mean-square errors of 0.751 and 0.781 for temperature and humidity, respectively.
In recent machine learning approaches, the evaluation of the model has often been limited to the overall prediction accuracy, and the data from all temperature ranges are treated equally, so the significant impact of extreme temperature on crops was not considered. Further, the models in prior works often performed badly in extreme conditions. The data used to train the models have often been collected from historical greenhouse data, within which the amount of extreme data is typically smaller than the amount of normal-temperature data. This lowers the prediction accuracy at extreme temperatures, increasing the probability that extreme temperatures may be misjudged as normal temperatures, with a higher likelihood of severe economic losses.
To address the above problems, in this paper, we propose an improved loss function. In contrast to the traditional mean squared error, the proposed improved loss function assigns different weights to extreme and normal temperatures, which can help models focus more on the extreme cases. The improved loss function calculates differently for predicted values higher or lower than the actual value. Therefore, models with this function are updated based on the types of wrong predictions in extreme temperatures, thereby improving the performance of models under extreme conditions.

The remainder of this paper is organized as follows. In section 2, the greenhouse dataset we used and preprocessing steps are introduced, along with a presentation of the theoretical basis of the work and the proposed algorithm. In section 3, we describe the experimental setup and discuss the experimental results. And section 4 provides conclusion for this work.

\section{Materials and Methods} \label{methods}

\subsection{DATA Description}
\begin{figure}[htp]
\vspace{0.5em}
\setlength{\belowcaptionskip}{-0.5cm}
\begin{center}
\includegraphics[width=1\linewidth]{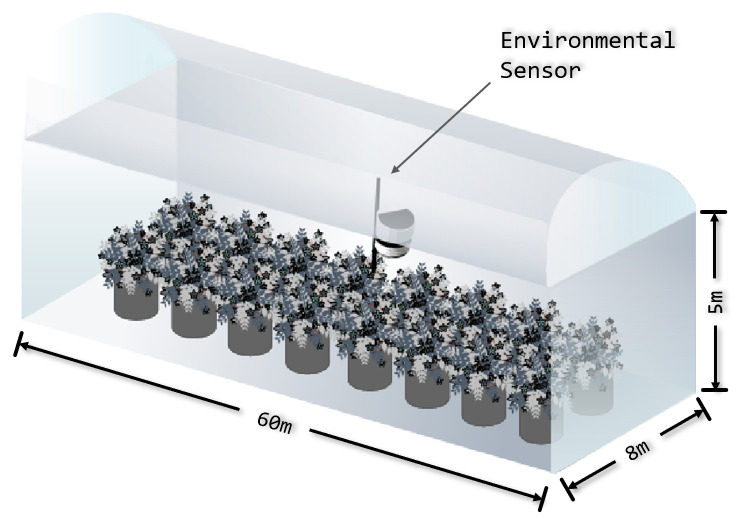}
\end{center}
\vspace{-1em}
   \caption{\textbf{Layout of the experimental greenhouse.} The environmental sensor is installed at a height of 1m.}
\label{layout}
\vspace{1em}
\end{figure}

Our data were collected from the solar greenhouse of the Zhuozhou Farm of China Agricultural University (39° 28' 8.19740'' N, 115° 51' 9.31366'' E). The simulated greenhouse is shown in Figure \ref{layout}. The greenhouse spans 60 m from east to west and 8 m from north to south and has a cold-formed steel frame structure. A suite of installed sensors collects environmental data including temperature, humidity, pressure, illumination, and carbon dioxide concentration at a frequency of one time per minute.

We collected approximately 14 months of data from June 17, 2018, to August 12, 2019, with approximately 410,000 valid data records. An example of the data samples is shown in Table \ref{table:1}.

\begin{table*}
\setlength{\belowcaptionskip}{0.25cm}
\caption{\textbf{Examples of data samples.}}
\label{table:1}
\vskip 0.15in
\vspace{-0.7cm}
\begin{center}
\begin{small}
\begin{tabular}{cccccc}
\toprule[1.2pt]
Time         & Temperature(°C)      & Humidity(\%rh)  & Pressure($10^2$ Pa)    & Illumination(lx)    & $CO_2$(ppm)   \\
\midrule
2018/06/17 19:35:23 & 27.90 & 60.8 & 997.9 & 657.0 & 611.0\\
2018/06/17 19:36:23	&27.90	&60.8	&997.9	&657.0	&611.0\\
2018/06/17 19:37:22	&27.90	&60.8	&997.9	&657.0	&611.0\\
2018/06/17 19:38:22	&27.90	&60.8	&998.0	&657.0	&611.0\\
2018/06/17 19:39:22	&27.90	&60.8	&998.0	&657.0	&611.0\\
2018/06/17 19:40:22	&27.90	&60.8	&998.0	&657.0	&611.0\\
…… &	……&	……&	……&	……&	……\\
\bottomrule[1.2pt]
\end{tabular}
\end{small}
\end{center}
\vskip -0.1in
\vspace{-1.5em}
\end{table*}

In Table \ref{table:1}, Temperature (°C) represents the temperature value inside the greenhouse, and its minimum accuracy was 0.01°C. Humidity (\%rh) denotes the relative humidity value inside the greenhouse, with a minimum accuracy of 0.1\% Pressure ($10^2$ Pa), illumination (lx), and CO2 (ppm) denote the air pressure, light intensity, and carbon dioxide concentration inside the greenhouse.

\subsection{Data Preprocessing}
\begin{figure*}[htbp]
\vspace{0.5em}
\setlength{\belowcaptionskip}{-0.5cm}
\begin{center}
\includegraphics[width=1\linewidth]{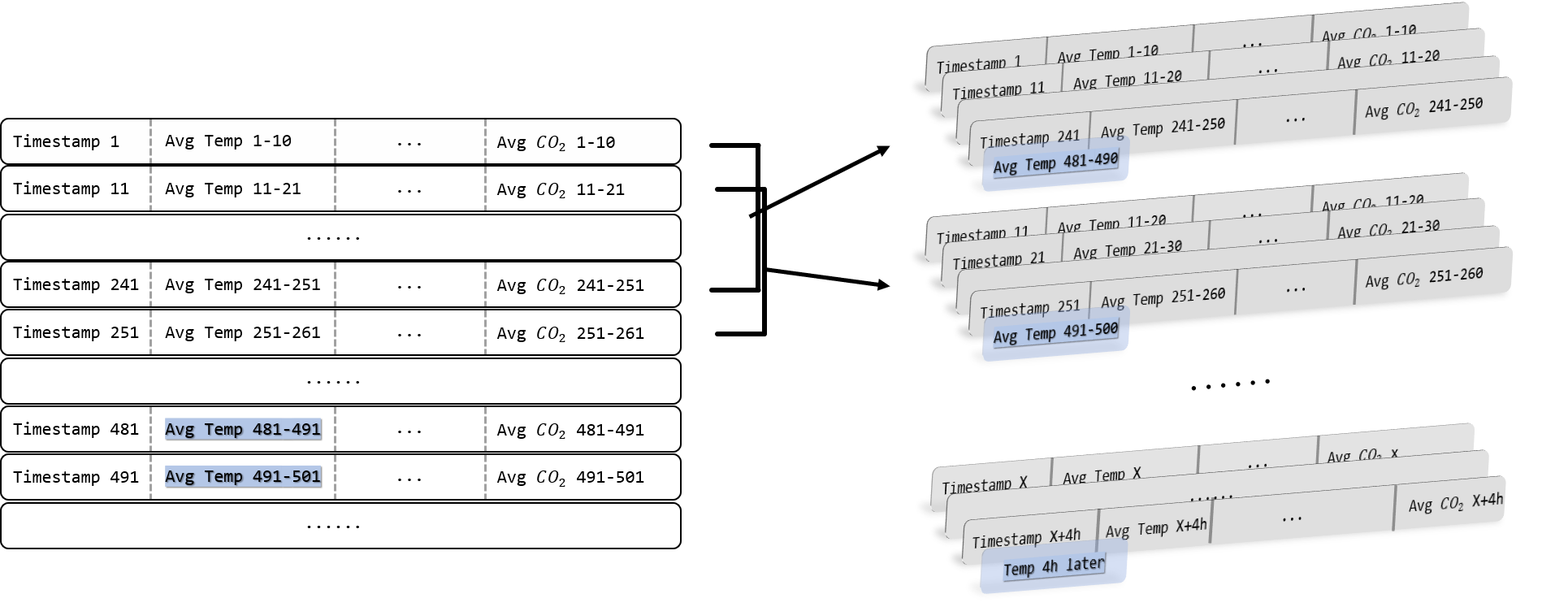}
\end{center}
\vspace{-1em}
   \caption{\textbf{Data reconstruction.} We used 4h of data to predict temperature 4h later.}
\label{reconstruction}
\vspace{1em}
\end{figure*}

We first perform data preprocessing before feeding them into machine learning algorithms. In this study, data preprocessing can be divided into four steps, including data cleaning, data aggregation, data reconstruction, and training/testing sets division.
Since greenhouse sensors may be unstable during data acquisition, anomalous or missing values may exist in raw data. Therefore, in data cleaning process, we remove data with duration less than 30 minutes, and then remove outliers whose temperature is more than 40°C or below than 0°C.
Compared with the changing rate of the greenhouse environment, the data collection frequency of one time per minute was relatively rapid. Thus, it was necessary to reduce the data sampling frequency before training the model. We average the collected greenhouse data at 10-minutes intervals. This strategy not only served to reduce the data frequency, but also to balance numerical fluctuations over short periods potentially caused by unstable sensors.
Since an early warning of extreme temperatures four hours in advance is sufficient for farmers to perform timely emergency remedial measures, we used four hours of data to predict temperature four hours in advance. We reconstruct the data to achieve this goal, as shown in Figure \ref{reconstruction}.

\begin{figure}[htp]
\vspace{-0.5em}
\setlength{\belowcaptionskip}{-0.5cm}
\begin{center}
\includegraphics[width=1\linewidth]{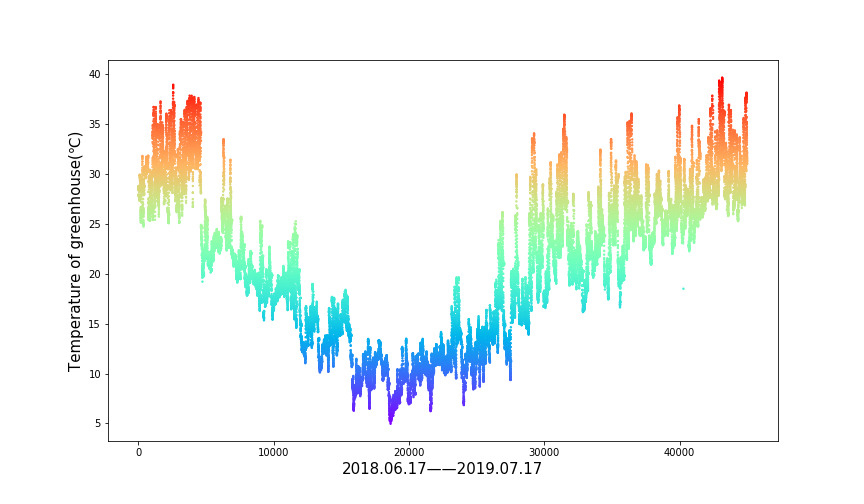}
\end{center}
\vspace{-1em}
   \caption{\textbf{Variation of solar greenhouse temperatures between 06/17/2018 and 7/17/2019.}}
\label{temperature}
\vspace{1em}
\end{figure}

Then, to better verify the performance of the model under extreme temperature conditions, we selected data from August 2018 (summer) and February 2019 (winter) as testing set. The size of the testing set is 6570 and that of the training set is 36244.
After the above process, the changing curve of greenhouse temperature from 2018 to 2019 is shown in Figure \ref{temperature}. When the temperature is higher than 30 °C, greenhouse crops are susceptible to heat damage \cite{Bita2013PlantTT, Chen1982AdaptabilityOC}. Plant chlorophyll can lose activity easily, and the dark reaction of photosynthesis can also be blocked, causing damage to plants \cite{Fahad2017CropPU}. Therefore, we define the data with temperature higher than 30°C as extremely high temperatures. Similarly, since crops are easily affected by cold damage below 10°C \cite{Huner2004PhotosynthesisPA,Liu2013TheLT}, we define data with temperatures less than 10°C as extremely low temperatures.

\subsection{MACHINE LEARNING BASICS IN EXTREME TEMPERATURES}
The mathematical model of the conventional machine learning algorithm used to predict temperature can be expressed as the following equation, where $x$ is the greenhouse environment data, $\theta$ is the parameter of the machine model, and $\hat{y}$ is the predicted future greenhouse temperature.

$\hat{y}=h(x,\theta)$

To maximization the accuracy of the prediction results of machine learning algorithms, a loss function must be introduced to measure the deviation of the predicted values from the ground truth. To obtain prediction results as close to the ground truth as possible, it was necessary to solve the following mathematical optimization problem on the training set.

$$\argmin_{\theta}\sum J(h(x_{i}, \theta), y_{i})$$

where $J$ is the loss function and $y_i$ is the true value to be predicted in the training set.

The learning process of the conventional machine learning algorithms is to solve the above formula iteratively by using specific optimization algorithms. Commonly used optimization algorithms include Adam \cite{Kingma2015AdamAM}, RMSprop \cite{tieleman2012lecture}, and Adagrad \cite{Duchi2011AdaptiveSM}. These optimization algorithms are based on the gradient descent algorithm, so they need to obtain the gradient of the prediction result on the training set, namely
$$
\operatorname{Grad}(\theta)=\sum \frac{\partial J\left(h\left(x_{i}, \theta\right), y_{i}\right)}{\partial \theta_{k}} \cdot \overrightarrow{\theta_{k}}
$$

according to the chain derivation rule

$$
\frac{\partial J\left(h\left(x_{i}, \theta\right), y_{i}\right)}{\partial \theta_{k}}=\frac{\partial J\left(h\left(x_{i}, \theta\right), y_{i}\right)}{\partial h(x, \theta)} \cdot \frac{\partial h(x, \theta)}{\partial \theta_{k}}
$$

Therefore,

$$
\operatorname{Grad}(\theta)=\frac{\partial J\left(h\left(x_{i}, \theta\right), y_{i}\right)}{\partial h(x, \theta)} \cdot \sum \frac{\partial h(x, \theta)}{\partial \theta_{k}} \cdot \overrightarrow{\theta_{k}}
$$

Therefore, the loss function has a decisive influence on the learning process of algorithms.

\subsection{IMPROVED LOSS FUNCTION}
Considering the temperature prediction problem, the predicted results can be classified into four category, as shown in Table \ref{table:2}.

\begin{table}[htbp]
\setlength{\abovecaptionskip}{0.25cm}
\setlength{\belowcaptionskip}{0.25cm}
\caption{\textbf{Classification of prediction results.}}
\label{table:2}
\vskip 0.15in
\vspace{-0.7cm}
\begin{center}
\begin{small}
\begin{tabular}{ccc}
\toprule[1.2pt]
Case &Predicted result &Ground truth\\
\midrule
TP	&Extreme	&Extreme\\
TN	&Normal	&Normal\\
FP	&Extreme	&Normal\\
FN	&Normal	&Extreme\\
\bottomrule[1.2pt]
\end{tabular}
\end{small}
\end{center}
\vspace{-1em}
\end{table}

In the false positive (FP) case, the prediction accuracy has little effect on production, whereas false negative (FN) prediction can potentially cause significant economic losses in the greenhouse, since it mispredict the extreme temperature as normal one. However, the most commonly used loss function for regression problems is the mean square error,

$$
J_{\mathrm{RSE}}\left(h\left(x_{i}, \theta\right), y_{i}\right)=\frac{1}{m} \sum_{i=1}^{m}\left(h\left(x_{i}, \theta\right)-\widehat{y}_{i}\right)^{2}
$$

It can be observed that the weight of each data sample is the same. However, the number of data samples of extreme cases was lower than that of normal cases; as a result, the performance of conventional machine learning models used in prior works may decline in extreme cases. Therefore, we propose an improved loss function to enhance prediction accuracy for extreme cases and reduce FN cases. 
We first increase the weight for extreme temperature samples during the training phase. The value of the weight increased was calculated by dividing the number of extreme cases and normal cases, as given below.

$$
\begin{aligned}
w_{H} &=\frac{\left|U_{\text {normal}}\right|}{\left|U_{\text {high}}\right|}, \\
w_{L} &=\frac{\left|U_{\text {normal}}\right|}{\left|U_{\text {low}}\right|}
\end{aligned}
$$

where $U_{low}=\{(x_i,y_i) | y_i\leq T_{low}, (x_i,y_i)\in U\}$, $U_{normal}=\{(x_i,y_i) | T_{low}\textless y_i\textless T_{high}, (x_i,y_i)\in U\}$, 
$U_{high}=\{(x_i,y_i) | y_i\geq T_{high}, (x_i,y_i)\in U\}$,and $U$ is the training set.
In addition, to reduce the occurrence of FN cases, we defined an extreme case importance factor $a$. In extreme cases, we increase the weight of data whose predicted result is within the real value and correspondingly decrease the weight of other data. Since machine learning algorithms with backpropagation mechanisms or leaf splitting strategies tend to pay more attention to data with larger weight, the introduction of the factor $a$ can make the prediction results of the model in extreme cases be more accurate.
In summary, the improved loss function definition $J(\theta)$ is as follows:

$$
J(\theta)=J_{\text {high}}(\theta)+J_{\text {low}}(\theta)+J_{\text {normal}}(\theta)
$$

where
$$
J_{\text {high}}(\theta)=\sum_{\left(x_{i}, y_{i}\right) \in U_{\text {high}}} w_{H} * f(x) *\left(h_{\theta}\left(x_{i}\right)-y_{i}\right)^{2} 
$$
$$
J_{\text {low}}(\theta)=\sum_{\left(x_{i}, y_{i}\right) \in U_{\text {low}}} w_{L} * g(x) *\left(h_{\theta}\left(x_{i}\right)-y_{i}\right)^{2} 
$$
$$
J_{\text {normal}}(\theta)=\sum_{\left(x_{i}, y_{i}\right) \in U_{\text {normal}}}\left(h_{\theta}\left(x_{i}\right)-y_{i}\right)^{2}
$$

Here, $f$ and $g$ are hinge functions:

$$
f\left(x_{i}\right)=\left\{\begin{array}{r}
a, y_{i}<h_{\theta}\left(x_{i}\right) \\
1-a, y_{i} \geq h_{\theta}\left(x_{i}\right)
\end{array},\left(x_{i}, y_{i}\right) \in U_{\text {high}}\right. 
$$
$$
g\left(x_{i}\right)=\left\{\begin{array}{r}
1-a, y_{i}<h_{\theta}\left(x_{i}\right) \\
a, y_{i} \geq h_{\theta}\left(x_{i}\right)
\end{array},\left(x_{i}, y_{i}\right) \in U_{\text {low}}\right.
$$

And the calculation process can be summarized by the following pseudo code.

\begin{algorithm}
   \caption{Improved loss function calculation process}
   \label{alg:example}
\begin{algorithmic}
   \STATE {\bfseries Input:} $\hat{y}=\{\hat{y_1},...,\hat{y_N}\}$, $a$, $T_H$, $T_L$\\
	\quad\qquad $\hat{y}$ is the predicted value of temperature \\
	\quad\qquad $y$ is the true value of temperature \\
	\quad\qquad $a$ is the extreme temperature importance factor\\
	\quad\qquad $T_H$ is the high temperature threshold\\
	\quad\qquad $T_L$ is the low temperature threshold\\
	\STATE {\bfseries Output:} Total loss $L_{total}$ \\
	Initialize $L_{total}=0$
   \STATE $y_H\gets\{y_i | y_i\geq T_H\}$
   \STATE $y_{normal}\gets\{y_i | T_L \textless y_i\textless T_H\}$
   \STATE $y_L\gets\{y_i | y_i\leq T_L\}$
   \STATE $w_H=\nicefrac{y_{normal}}{y_H}$
   \STATE $w_L=\nicefrac{y_{normal}}{y_L}$
   
   \STATE {\bfseries for}\ each pair $\hat{y_i}$ and $y_i$\ {\bfseries do}:
    
   \STATE \quad\quad {\bfseries if} $y_i\geq T_H$:
   
   \STATE \quad\quad\quad\quad {\bfseries if} $y_i \textgreater  \hat{y_i}$:
   \STATE \quad\quad\quad\quad\quad\quad $L\gets w_H*a*(y_i-\hat{y_i})^2$
   \STATE \quad\quad\quad\quad {\bfseries else}:
   \STATE \quad\quad\quad\quad\quad\quad $L\gets w_H*(1-a)*(y_i-\hat{y_i})^2$
   \STATE \quad\quad\quad\quad {\bfseries end if}
   
   \STATE \quad\quad {\bfseries else if} $y_i\leq T_L$:
   
    \STATE \quad\quad\quad\quad {\bfseries if} $y_i \textgreater\ \hat{y_i}$:
   \STATE \quad\quad\quad\quad\quad\quad $L\gets w_L*(1-a)*(y_i-\hat{y_i})^2$
   \STATE \quad\quad\quad\quad {\bfseries else}:
   \STATE \quad\quad\quad\quad\quad\quad $L\gets w_L*a*(y_i-\hat{y_i})^2$
   \STATE \quad\quad\quad\quad {\bfseries end if}
   
   \STATE \quad\quad {\bfseries else}:
   \STATE \quad\quad\quad\quad $L\gets (y_i-\hat{y_i})^2$

    \STATE \quad\quad{\bfseries end if}
    \STATE \quad\quad $L_{total}\gets L_{total}+L$
    
   \STATE {\bfseries end for}
   \STATE {\bfseries return} $L_{total}$

\end{algorithmic}
\end{algorithm}

\section{EXPERIMENTS AND RESULTS}

\subsection{EXPERIMENTAL SETUP}
We conduct experiments on our preprocessed dataset. Data from August 2018 and February 2019, the number of which is 6,570, are used for testing, and the remaining 36244 data are used for training. And they are conducted on a computer workstation using the Windows 10 operating system with an Intel Core i9-9900K CPU at 3.60 GHz, with 16 GB of memory installed as well as an Nvidia GeForce RTX 2080 Ti graphics processing unit (11GB), and all models are implemented using the Python3 programming language.
To verify the versatility of the improved loss function, we conduct comparative experiments on the basic models of LightGBM (LGB) \cite{ke2017lightgbm}, long short-term memory (LSTM) \cite{hochreiter1997long}, and artificial neural network (ANN) \cite{jain1996artificial}, and evaluate the models after applying the improved loss function. Furthermore, to further verify the optimization effect of the improved loss function, we select a = 0.5, 0.7, and 0.9 for comparison and compare the basic LGB model to the LGB with improved loss function under extreme temperature conditions.
When comparing models with different extreme case important factors, we use the same hyperparameters. The BPNN's hyperparameters are shown in Table \ref{table:a1}. We used PyTorch to build the BPNN model. We used a three-layer neural network (excepting the input layer), and the number of nodes was the same as \cite{Ahmad2014NeuralNM}.

\begin{table}[htbp]
\setlength{\belowcaptionskip}{0.25cm}
\caption{\textbf{Settings of BPNN hyperparameters.}}
\label{table:a1}
\vskip 0.15in
\vspace{-0.7cm}
\begin{center}
\begin{small}
\begin{tabular}{cc}
\toprule[1.2pt]
Algorithm&	BPNN\\
Number of hidden layers&	3\\
Number of nodes in each hidden layer&	6, 3, 1\\
Optimizer&	Adam\\
Learning rate&	0.001\\
Batch size&	2000\\
Epoch&	100\\

\bottomrule[1.2pt]
\end{tabular}
\end{small}
\end{center}
\end{table}

The hyperparameter settings used for the LSTM are shown in Table \ref{table:a2}. We also use PyTorch to build the LSTM model.

\begin{table}[htbp]
\setlength{\belowcaptionskip}{0.25cm}
\caption{\textbf{Settings of LSTM hyperparameters.}}
\label{table:a2}
\vskip 0.15in
\vspace{-0.7cm}
\begin{center}
\begin{small}
\begin{tabular}{cc}
\toprule[1.2pt]
Algorithm&	LSTM\\
Number of LSTM layers&	1\\
Number of nodes in LSTM layer&	48\\
Optimizer&	Adam\\
Learning rate&	0.02\\
Batch size&	2000\\
Epoch&	100\\

\bottomrule[1.2pt]
\end{tabular}
\end{small}
\end{center}
\end{table}

The key hyperparameter settings used for the LGB model are shown in Table \ref{table:a3}. We use python library LightGBM to implement LGB model.

\begin{table}[htbp]
\setlength{\belowcaptionskip}{0.25cm}
\caption{\textbf{Settings of LSTM hyperparameters.}}
\label{table:a3}
\vskip 0.15in
\vspace{-0.7cm}
\begin{center}
\begin{small}
\begin{tabular}{cc}
\toprule[1.2pt]
Algorithm&	LGB\\
Boosting type&	Gbdt\\
Learning rate&	0.05\\
Epoch&	100\\

\bottomrule[1.2pt]
\end{tabular}
\end{small}
\end{center}
\end{table}

\subsection{MODEL PERFORMANCE EVALUATION}
To effectively evaluate the predictive performance of the model, the evaluation indicators used in this paper include mean absolute error (MAE), mean bias and recall rate. The evaluation index calculation formula is as follows:
$$
\text {MAE}=\frac{1}{m} \sum_{i=1}^{m}\left|y_{i}-\hat{y}_{l}\right| 
$$
$$
\text {Mean bias}=\frac{1}{m} \sum_{i=1}^{m}\left(y_{i}-\widehat{y}_{l}\right) 
$$
$$
\text{Recall}=\frac{TP}{TP+FN}
$$
The definitions of TP, FP, and FN are the same as those given in Table \ref{table:2}. The higher the recall rate, the lower the FN, the more extreme cases were correctly predicted.
To test the effect of improved loss function, we first examine the performance of the three basic models, then examine the performance of a model that only increase the weight of extreme cases ($a=0.5$), and finally test the performance of a model with $a=0.9$.

\begin{table*}[htbp]
\setlength{\abovecaptionskip}{0.15cm}
\setlength{\belowcaptionskip}{0.25cm}
\caption{\textbf{Predicted results in test set.}}
\label{table:3}
\vskip 0.15in
\vspace{-0.7cm}
\begin{center}
\begin{small}
\begin{tabular}{cccccc}
\toprule[1.2pt]
Model&	Improved loss function&	Recall&	MAE (high)&	MAE (low)&	MAE (normal)   \\
\midrule
\multirow{3}{*}{LGB} &	\xmark&	0.82&	2.15&	0.79&	0.67\\
	&a=0.5&	0.87&	1.71&	0.64&	0.78\\
	&a=0.9&	0.87&	1.56&	0.59&	0.85\\
\multirow{3}{*}{LSTM}&	\xmark&	0.68&	2.31&	1.90&	1.06\\
	&a=0.5&	0.75&	1.61&	1.53&	1.57\\
	&a=0.9&	0.78&	1.60&	1.33&	1.96\\
\multirow{3}{*}{ANN}	&\xmark&	0.78&	1.98&	1.11&	0.98\\
	&a=0.5&	0.82&	1.72&	0.84&	1.62\\
	&a=0.9&	0.83&	1.44&	1.11&	1.91\\

\bottomrule[1.2pt]
\end{tabular}
\end{small}
\end{center}
\vskip -0.1in
\vspace{-1em}
\end{table*}

As shown in Table \ref{table:3}, the MAE of the basic models in extreme cases is higher than the normal case, which demonstrates that models do have performance degradation in extreme temperature. With the introduction of the high- and low-temperature weighting factors $w_H$ and $w_L$, all three models ($a=0.5$) obtain a lower MAE than the basic models for both the high and low temperatures. For LSTM, when $a=0.5$, the MAE value at high temperature decrease from 2.31 to 1.61, and the MAE value at low temperature decrease from 1.90 to 1.53. The recall rate of the models also improve, which shows that the proposed method reduce the FN cases successfully. Although the performance of the model under normal temperature reduce a little, the growth of the plants is almost unaffected when temperature fluctuate in the normal range.
When $a=0.9$, the results of three models have been further improved, which shows the function $f$ and $g$ perform well in all models, which can reduce the FN cases and reduce the MAE under extreme temperatures.
To further determine the influence of the extreme case importance factor on the model, we take LGB model as an example and further compare the prediction results with different $a$. The results are shown in Table \ref{table:4}.

\begin{table}[htbp]
\setlength{\belowcaptionskip}{0.25cm}
\caption{\textbf{LGB comparison results.} We calculate the mean bias separately on high temperature and low temperature. When the mean bias is less than zero, the predicted value is less than the ground truth, and vice versa.}
\label{table:4}
\vskip 0.15in
\vspace{-0.7cm}
\begin{center}
\begin{small}
\begin{tabular}{ccc}
\toprule[1.2pt]
Improved loss function&	Mean bias (high)&Mean bias (low)\\
\midrule
\xmark	&-1.95&	0.72\\
$a=0.5$&	-1.39&	0.45\\
$a=0.7$&	-1.26&	0.33\\
$a=0.9$&	-1.08&	0.28\\

\bottomrule[1.2pt]
\end{tabular}
\end{small}
\end{center}
\vskip -0.1in
\vspace{-1.5em}
\end{table}

The results in Table \ref{table:4} show that when the extreme case importance factor a change from 0.5 to 0.9, the performance of the improved LGB under extreme temperatures gradually increase. When $a=0.9$, the mean bias under high-temperature conditions increase from \texttt{ground truth-1.95°C} to \texttt{ground\ truth-1.08°C}, which means the possibility of incorrectly predicting high temperature as normal temperature is reduced. Similarly, under low-temperature conditions, the mean bias is reduced from \texttt{ground\ truth+0.72°C} to \texttt{ground\ truth+0.28°C}, indicating that the improved loss function also function well on low temperature. 

\begin{figure}[htbp]
\centering
\subfigure[]{
\includegraphics[width=9cm]{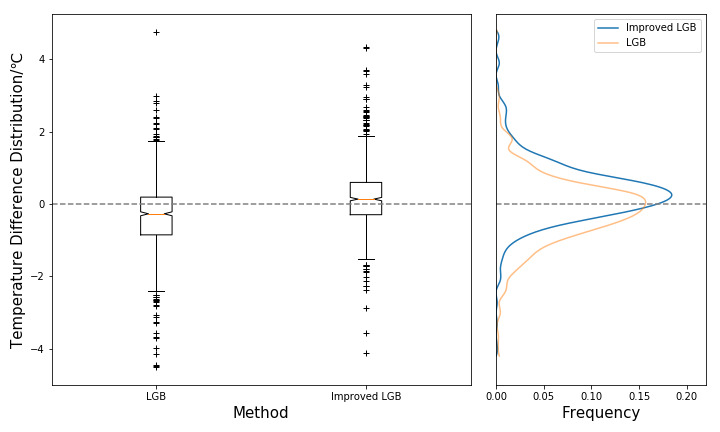}
\vspace{-0.5cm}
}
\quad
\subfigure[]{
\includegraphics[width=9cm]{computer 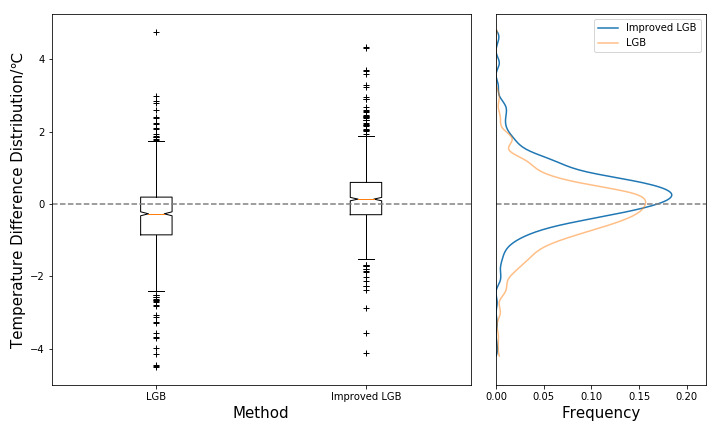}
\vspace{-0.5cm}
}
\caption{\textbf{Comparison of prediction results at high temperatures (a) and low temperatures (b).}}
\setlength{\abovecaptionskip}{-0.5cm}
\setlength{\belowcaptionskip}{-0.5cm}
\vspace{-0.5em}
\end{figure}

To show the improvement in the prediction results, we use a box plot to compare the basic model with the improved one and visually show the temperature difference distribution, as exhibited in Figure 4. Under high temperature, the median value of temperature difference increases from $-0.26$ to $0.13$, the mean value increases from $-0.32$ to $0.19$, and the standard deviation reduces from $0.9$ to $0.86$. Under low temperature, the mean value decrease from $0.28$ to $-0.12$, the median value reduces from $0.20$ to $-0.15$, and the standard deviation reduces from $0.57$ to $0.49$. This means that our approach can reduce the standard deviation and can shift the prediction results towards extreme temperatures, reducing the possibility of missed prediction in extreme cases.

\section{Conclusions} \label{Conclusions}
In order to forecast the freezing or heat damage that crops may suffer, it is essential to predict extreme temperatures accurately and timely. In this paper, an improved loss function that can be applied to various machine learning algorithms was proposed. The function mainly includes two parts, which are extreme weight and adjustment function. Then, we collected 14 months of various greenhouse sensor data from Zhuozhou Farm’s solar greenhouse and perform data cleaning, data aggregation, data reconstruction, and training/testing sets division sequentially. By implementing the improved loss function to LGB, LSTM, and ANN, the results show that compared with the basic models, the performance of the improved model becomes better in extreme cases. Specifically, the recall rate of the prediction results under extreme conditions increases significantly and the MAE value decreases at the same time. It proves that the proposed approach can better guarantee timely judgment of extreme temperatures in greenhouses, thereby preventing unnecessary losses caused by incorrect and missed predictions.

\ack{
This research did not receive any specific grant from funding agencies in the public, commercial, or not-for-profit sectors.
}

\nocite{*}


{\small
\normalem
\bibliographystyle{compj}
\bibliography{egbib}
}

\end{document}